\documentclass[runningheads]{llncs}
\usepackage[dvipdfmx]{graphicx}
\usepackage{color}
\newcommand{\letter}[1]{`{\tt #1}'}
\newcommand{\letterp}[1]{`{\tt #1}.'}
\newcommand{\letterc}[1]{`{\tt #1},'}   
\newcommand{\word}[1]{``{\tt #1}''}
\newcommand{\wordp}[1]{``{\tt #1}.''}
\newcommand{\wordc}[1]{``{\tt #1},''}

\makeatletter
\newcommand{\figcaption}[1]{\def\@captype{figure}\caption{#1}}
\newcommand{\tblcaption}[1]{\def\@captype{table}\caption{#1}}
\makeatother
\begin{document}
\title{Learning to Kern: Set-wise Estimation of Optimal Letter Space}
\titlerunning{Learning to Kern}
%

\author{Kei Nakatsuru\and
Seiichi Uchida\orcidID{0000-0001-8592-7566}}
\authorrunning{K. Nakatsuru and S. Uchida}

\institute{Kyushu University, Fukuoka, Japan\\
\email{uchida@ait.kyushu-u.ac.jp}}

\maketitle              
\begin{abstract} 
Kerning is the task of setting appropriate horizontal spaces for all possible letter pairs of a certain font. One of the difficulties of kerning is that the appropriate space differs for each letter pair. Therefore, for a total of 52 capital and small letters, we need to adjust $52 \times 52 = 2704$ different spaces. Another difficulty is that there is neither a general procedure nor criterion for automatic kerning; therefore, kerning is still done manually or with heuristics. In this paper, we tackle kerning by proposing two machine-learning models, called pairwise and set-wise models. The former is a simple deep neural network that estimates the letter space for two given letter images. In contrast, the latter is a transformer-based model that estimates the letter spaces for three or more given letter images. For example, the set-wise model simultaneously estimates 2704 spaces for 52 letter images for a certain font. Among the two models, the set-wise model is not only more efficient but also more accurate because its internal self-attention mechanism allows for more consistent kerning for all letters. Experimental results on about 2500 Google fonts and their quantitative and qualitative analyses show that the set-wise model has an average estimation error of only about 5.3 pixels when the average letter space of all fonts and letter pairs is about 115 pixels.

\end{abstract}
\begin{figure}[t]
    \centering
    \includegraphics[width=.7\textwidth]{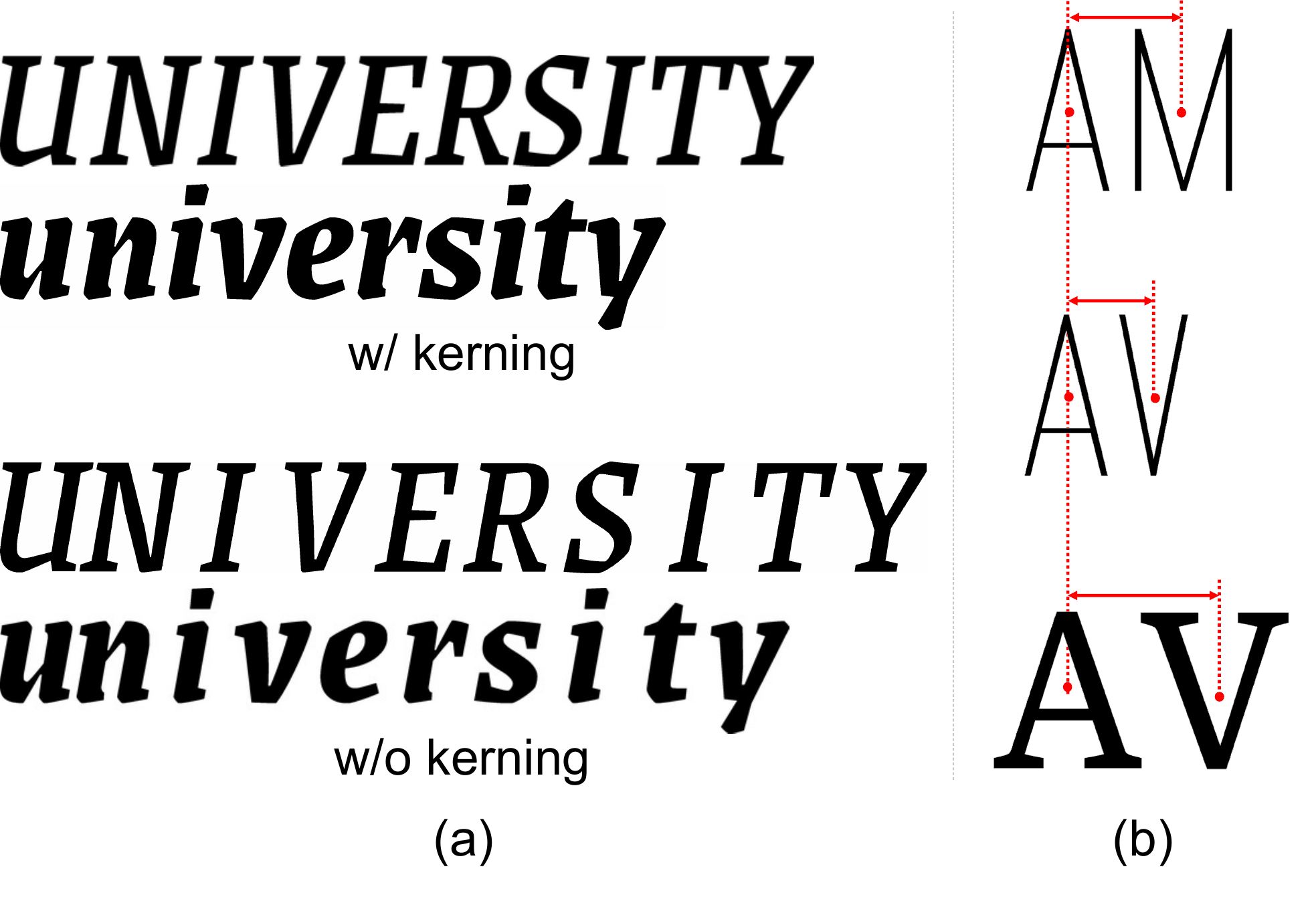}\\[-3mm]
    \caption{(a)~Effect of kerning. (b)~Letter spaces vary with font styles and letter pairs. This paper measures the letter space by the horizontal distance between the centers of adjacent letters (marked by red dots).}
    \label{fig:kerning}
\end{figure}

\section{Introduction\label{sec:intro}}
{\em Kerning}, or letter spacing, is the micro-typographic task of adjusting the horizontal space between adjacent letters\cite{Kindersley}. For monospaced fonts (like most Chinese fonts), the space is easily determined so that all letters occupy the {\em same} horizontal space. In contrast, kerning is a very important operation for the proportional fonts of the Latin alphabet letters. Fig.~\ref{fig:kerning}(a) shows the effect of kerning. Without kerning (that is, by monospacing), the letters do not appear to form the word \wordc{UNIVERSITY} especially due to an excessive space around \letterp{I} 
Various cognitive psychology research on reading has revealed that kerning is important for better readability~\cite{arditi1990,chung2002}. Therefore, kerning is not a trivial task for type designers and is beneficial for readers.\par
A careful observation of Fig.~\ref{fig:kerning}(a) proves that kerning depends on letters and letter pairs. For example, the spaces around \letter{I} become narrower from the monospaced case, whereas the space between \letter{U} and \letter{N} becomes wider. The space between \letter{E} and \letter{R} is almost the same as the monospaced case. The same letter class dependency is found in 
\wordp{university}  As shown in Fig.~\ref{fig:kerning}(b), the letter space also depends on font styles, as well as letter pairs.\par
Currently, kerning is determined manually or by heuristic rules.
For the manual cases, kerning is an enormous task; for the 52 letters of the Latin alphabet (i.e., \letter{A} to \letter{z}), type designers need to determine the space for each of $52^2=2704$ possible letter pairs. To relax this task, heuristic rules have also been used. A well-known heuristic is {\em optical spacing}~\cite{Hurst2009}, which tries to keep the blank space area between adjacent letters constant. (For example, the blank space between \letter{A} and \letter{V} in Fig.~\ref{fig:kerning}(b) is the parallelogram-like region between them. Its area is finally measured with additional perceptual heuristics~\cite{Karow1998}\footnote{The ``hz-program''~\cite{Karow1998,Han2000,Zapf1993} by a German type foundry called URW is a well-known optical spacing software not for letter pairs but for text string. About font design software, the following link is useful: \tt{ https://www.monotype.com/} \tt{resources/introduction-software-type-design}}.) 
It should be noted that heuristic rules are imperfect --- after applying the rules, a final adjustment process by some kerning specialists is still required. For example, although FontForge is one of {\it de-facto} standard font design software tools, which have a heuristics-based automatic kerning mode, users need to revise its kerning result.  
\par

\begin{figure}[t]
    \centering
    \includegraphics[width=.9\linewidth]{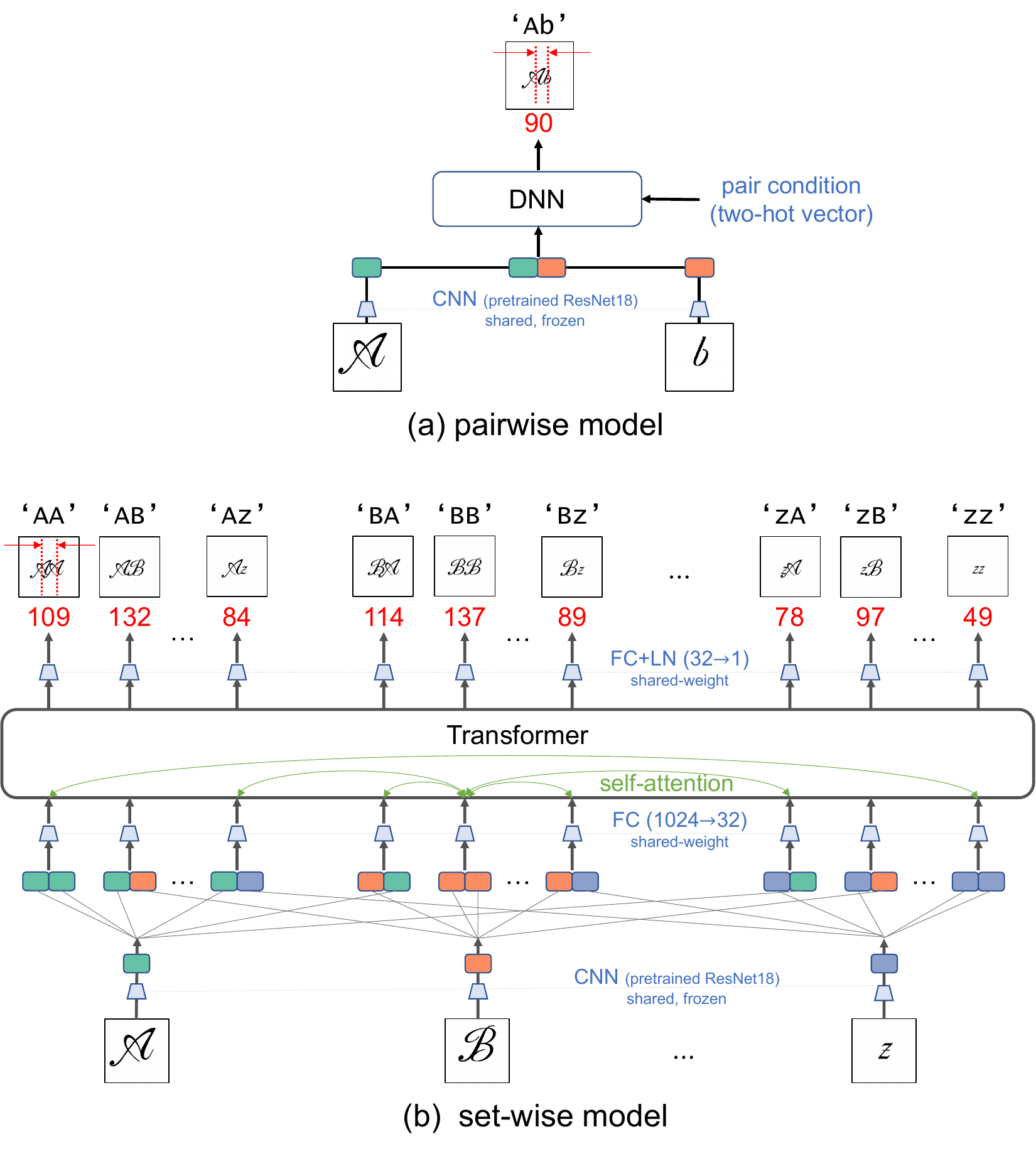}\\[-3mm]
    \caption{Overview of the proposed models for automatic letter-spacing: (a)~The pairwise model and (b)~the set-wise model.}
    \label{fig:overview}
\end{figure}

To the authors' knowledge, this paper is the first comprehensive attempt to estimate appropriate spaces for letter pairs by using machine-learning models. If the models are trained appropriately with existing font sets while using their manually-adjusted spaces as the ground-truth, they can automatically give the appropriate letter spaces of newly-designed fonts. Nowadays, everyone can design new fonts with software and font generation models, such as GANs and diffusion models. The proposed model will help to arrange the letters of the newly designed fonts for better readability. 
\par
We propose two models for learning to kern, that is, for estimating the optimal letter spacing. The simpler one is a {\em pairwise} model, which estimates the space between two given letter images, as shown in Fig.~\ref{fig:overview}(a). The pairwise model has two practical limitations. First, we must repeat the estimation $N^2$ times to determine the spaces of all $N^2$ pairs by $N$ letters. Second, it cannot utilize the other $N-2$ letters than given two letters during the estimation --- it does not coincide with the fact that type designers adjust spaces by looking at all $N$ letters in a font set for balanced and consistent kerning. For example, the space between \letter{H} and \letter{E} should be similar to the space between similar pairs, such as \letter{N} and \letterp{F} \par
The other model is a {\em set-wise} model, which estimates the $N^2$ spaces simultaneously, as shown in Fig.~\ref{fig:overview}(b). This model is based on Transformer~\cite{vaswani2017}. From $N$ letter images,  $N^2$ inputs are prepared and fed to Transformer, and $N^2$ estimated letter spaces are output from the Transformer. The set-wise model has two advantages over the pairwise model. First, the set-wise model estimates all $N^2$ spaces simultaneously. Second and more importantly, the set-wise model utilizes all $N$ letter images by its self-attention mechanism of Transformer for more accurate estimation. The set-wise model is thus expected to realize balanced and consistent kerning.\par
We perform various experimental evaluations quantitatively and qualitatively using capital and small letter images of 2558 fonts from Google Fonts. Since this is the first attempt at automatic kerning based on machine learning, there are no state-of-the-art models to be compared; we, therefore, compare the proposed two models with the {\it de-facto} standard automatic kerning by FontForge. The results show the advantages of using the proposed models, especially the set-wise model. 

\section{Related work} 
\subsection{Typographic theory of kerning}
As noted above, kerning is currently performed by manual inspection and/or heuristics, and there is no machine-learning model for automatic kerning. The literature \cite{lps50911995} advises that the spaces for \wordc{HH} \wordc{HO} and \word{OO} need to be first determined manually for the capital letters because \letter{H} has the representative straight lines on both sides and \letter{O} has curves. (For small letters, the literature \cite{lps50911995} recommends \letter{n} and \letter{o} as the first choices.) Then, the spaces of other pairs by using these spaces as references. Although this strategy is reasonable, we will encounter exceptional cases like italic fonts. Moreover, these pre-determined spaces are just references, and the spaces for other letters should be carefully adjusted individually while referring to them.  
\par
There are psychological research attempts to prove the importance of kerning. For example, several attempts~\cite{chung2002,galliussi2020,zorzi2012} have reported that increasing letter spaces improves the speed and accuracy of reading for dyslexia. Chung et al.~\cite{chung2002} also report that non-dyslexia persons often show similar improvement. On the other hand, other literature~\cite{paterson2010,terry1976,vinckier2011} reports that inappropriate letter spaces disturb word recognition ability. \par
As noted in Section~\ref{sec:intro}, the existing automatic kerning methods rely on heuristic rules. Heuristics-based methods, such as the automatic kerning function of FontForge\footnote{\url{https://fontforge.org/}}, need revisions by human experts; in other words, as we will see in the later experiments, letter spaces by the heuristics-based methods are still imperfect. Therefore, we propose machine learning-based automatic letter space estimation models, which are trained using various fonts and their default letter spaces by human experts\footnote{We find a non-reviewed report at {\tt https://project-archive.inf.ed.ac.uk/ug4/}\allowbreak{\tt 20170911/ug4\_proj.pdf}, which seems a project report by a master course student. It uses a simple regression model with a single peripheral feature (see Section~\ref{sec:peripheral} of our paper); namely, it is similar to a pair-wise model with the peripheral feature. As we will see later, our set-wise estimation model is better than the pairwise model, and our whole-image feature (by a CNN) is better than the peripheral feature.}.
\subsection{Automatic typography}
As we emphasized, kerning, which is a task of micro-typography, is a novel target of machine learning; in contrast, macro-typographic tasks, such as text layout, are tackled by recent machine learning methodologies. For example, Li et al.~\cite{li2020} generate a text layout by considering appropriate reading order as a constraint. We can find other layout generators that utilize Transformers or diffusion models~\cite{arroyo2021,chai2023,inoue2023,kong2022}. Nguyen et al.~\cite{nguyen2021} focus on the diversity of the generated layouts.

\section{Optimal Letter Space Estimation by Machine Learning}
This section details two models, pairwise and set-wise models, for optimal letter space estimation.

\subsection{The pairwise model\label{sec:pairwise}}
As shown in Fig.~\ref{fig:overview}(a), the pairwise model estimates the optimal letter space between given two letter images, say \letter{A} and \letterp{B} Among several techniques to accept a pair of images, we use vector concatenation after embedding each image into a 512-dimensional feature vector by ResNet18~\cite{he2016} 
pre-trained on the $N$-letter classification task. The 1024-dimensional concatenated vector is fed to a DNN with three fully-connected layers. The network is trained by Mean Absolute Error (MAE) between the estimated and ground-truth letter spaces.
\par
There are several possible ways to use the pairwise model for estimating the letter space of all the $N^2$ letter pairs. A straightforward way is to prepare $N^2$ different models for all pairs. Another way is to use a conditional DNN where the classes of the paired letters are fed to the DNN as its conditions. In this paper, we adopt the latter way because of its efficiency.
\subsection{The set-wise model\label{sec:setwise}}
Fig.~\ref{fig:overview}(b) shows the set-wise model that can estimate all $N^2$ letter spaces for $N$ letter classes. The pretrained ResNet18 first extracts the 512-dimensional feature vectors for each of the $N$ letter images of a certain font. Then, the feature vectors for a pair of two letters are concatenated as a 1024-dimensional vector and converted as a 32-dimensional vector by a single fully-connected layer (i.e., a linear transformation). All vectors of the $N^2$ class pairs are fed to a Transformer. Finally, the Transformer outputs 32-dimensional vectors, each of which is converted into the letter space between two letters by a single fully-connected layer with layer normalization. Like the pairwise model, MAE is also used as the loss function.\par
As mentioned in Section~\ref{sec:intro}, the set-wise model has two advantages that the pairwise model does not have. First, the set-wise model can accept all $N$ letter images and simultaneously estimate all $N^2$ letter spaces. Generally, for a newly designed font, letter spacing is necessary for all letter pairs. Hence, this property is a clear advantage of the set-wise model over the pairwise model, which requires $N^2$ iterations while changing class pairs.\par
The second and more important advantage of the set-wise model is that it can utilize the shapes of all $N$ letters during the letter space estimation. As noted in Section~\ref{sec:intro}, the set-wise model determines the letter space between \letter{H} and \letter{E} while watching the letter shapes of not only \letter{H} and \letter{E} but also the other letters, such as \letter{N} and \letterp{F} As type designers do, kerning is a set-wise optimization task for all $N^2$ letter spaces. Even if the space between two letters seems optimal for them, it will not be optimal by considering the balance and consistency with other letters. The well-known self-attention mechanism in Transformer naturally realizes a mutually dependent estimation of all $N^2$ letter spaces. 
\par
\section{Experimental Results}

\begin{figure}[t]
    \centering
    \includegraphics[width=.9\linewidth]{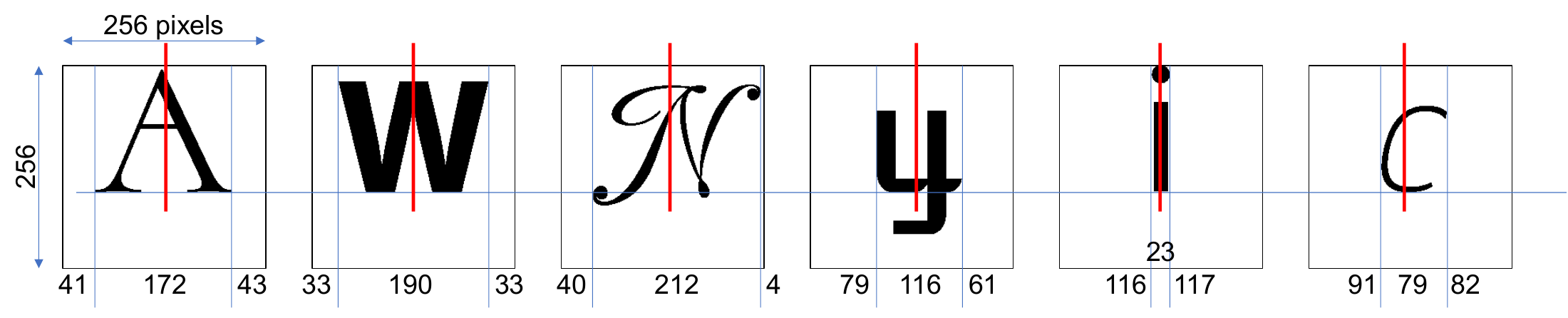}\\[-2mm]  
    \caption{Letter image examples. For example, the letter image \letter{A} has a width of 172 pixels, with left and right margins of 41 and 43 pixels, respectively. The red vertical line indicates the horizontal center of gravity of each letter.}
    \label{fig:image-example}  
  \end{figure}
  
 \begin{figure}[t]
    \centering   
    \includegraphics[width=.85\linewidth]{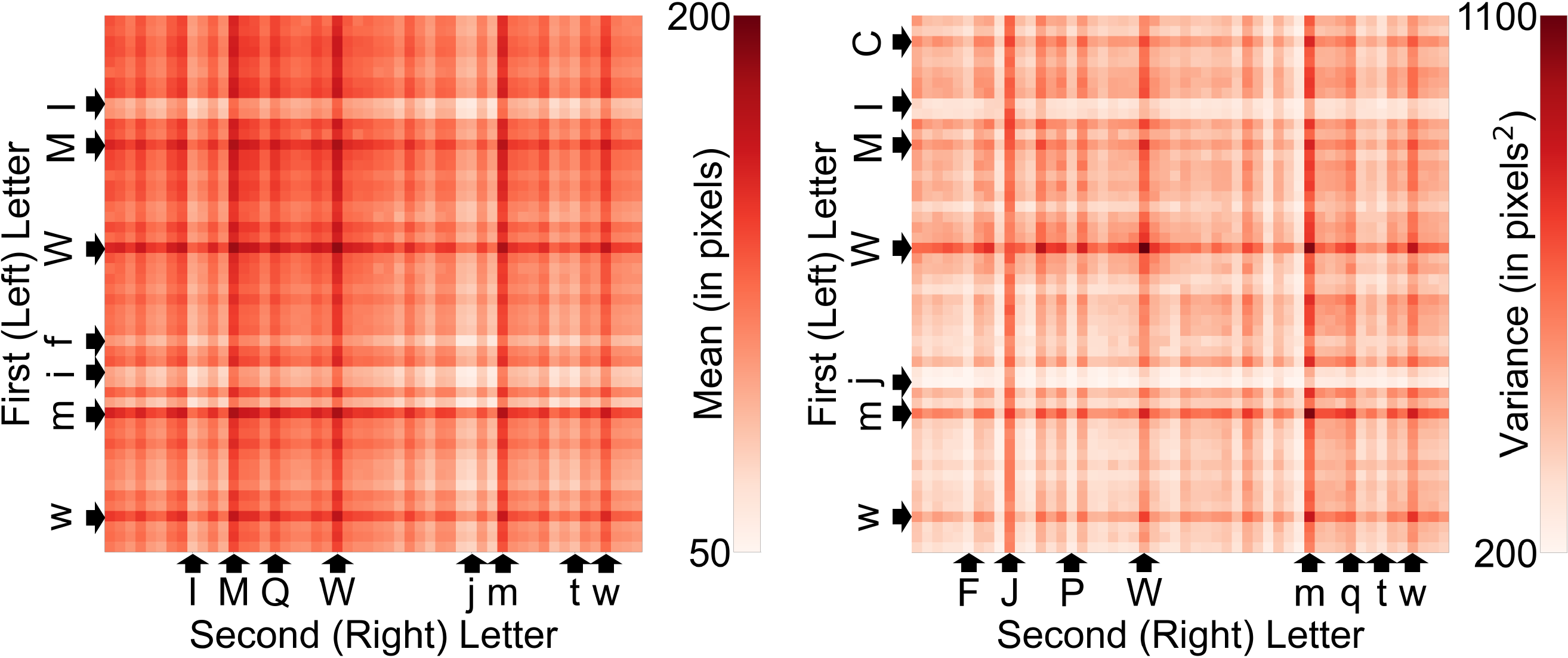}\\[-2mm]  
    \caption{Mean (left) and variance (right) of the ground-truth spaces of all letter pairs. The vertical and horizontal axes correspond to \letterc{A}$\cdots$\letterc{Z}\letterc{a}$\cdots$\letterp{z}}
    \label{fig:GT_Variance_withsmall}  
\end{figure}

\subsection{Dataset}
Google Fonts\footnote{\url{https://fonts.google.com/}} are used in our experiments.
After excluding monospaced fonts, 2558 Google Fonts are split into training, validation, and test sets with 2046, 256, and 256 fonts, respectively. To avoid data leakage, fonts in the same font family are not split into different sets. For example, the font family {\tt Grenze} is comprised of several fonts, such as {\tt Grenze Bold} and {\tt Grenze Extralight}, and all of them are used only in the test set. Each font is annotated with one of four styles: Serif, Sans-serif, Handwriting, and Display. The 2558 fonts comprise 626 Serif, 1283 Sans-serif, 200 Handwriting, and 449 Display fonts. From each font, we use 26 Latin alphabet capital letters (\letter{A}-\letter{Z}) and 26 small letters (\letter{a}-\letter{z}); therefore, $N=52$. \par
Each letter is rasterized as a 256$\times$256 binary image. Fig.~\ref{fig:image-example} shows several letter image examples. The baseline is set at a height of 96 pixels from the bottom (i.e., 160 pixels from the top) for all fonts. Then, the 52 letters of each font are magnified by the same scaling factor, with the constraint that they all fit into the $256\times 256$ region while staying on the baseline and preserving the aspect ratio. (The scale factor will differ depending on the font, but this is not a problem because the original proportions are maintained within the 52 letters of the font.)
\par
The default letter space (specified for each font) is used as the ground-truth. Specifically, as shown in Fig.~\ref{fig:kerning}(b), the letter space is measured as the horizontal pixel distance between the centers of letter images aligned by the default kerning rule of the font. Note that the center of a letter image is defined as the center of the gravity of black pixels. The average letter space of all letter pairs of all fonts is about 115 pixels.\par
Fig.~\ref{fig:GT_Variance_withsmall} shows the mean and variance of the ground-truth spaces of all $52\times 52$ letter pairs.
Variations in mean values and non-zero variance values prove that letter spaces differ for different letter pairs and different fonts, respectively, as already suggested in Fig.~\ref{fig:kerning}(b). These heatmaps also suggest that the letter pairs with larger mean values tend to have larger variance values. 
Several letter pairs, such as \word{W*}\footnote{The notation \word{W*} means arbitrary letter pairs where \letter{W} is the first letter.} and \word{*W}, show large variances. These large variances imply that estimating the spaces around \letter{W} is harder.
\subsection{Implementation details}
The network architectures of the pairwise and set-wise models are already outlined in Sections~\ref{sec:pairwise} and \ref{sec:setwise}, respectively. ResNet18 for extracting 512-dimensional feature vectors from individual letter images is pretrained by the 52-class classification task of all training letter images and then frozen. The number of Transformer layers  (i.e., Transformer blocks) is set at 1; in our preliminary experiment, the performance of the set-wise model is not largely affected by the number of Transformer layers. The number of self-attention heads is also set at 2 by a preliminary experiment. The feed-forward module in the Transformer has two layers. Both models use ReLU as the activation function. During training, we use Adam as the optimization algorithm with a batch size of 64. The standard early-stopping rule (no improvement on the validation loss for 100 epochs) determines the optimal training epochs. The learning rate is set to $10^{-4}$ for the pairwise model and $10^{-3}$ for the set-wise.\par
\begin{figure}[t]
    \centering
    \includegraphics[width=.8\linewidth]{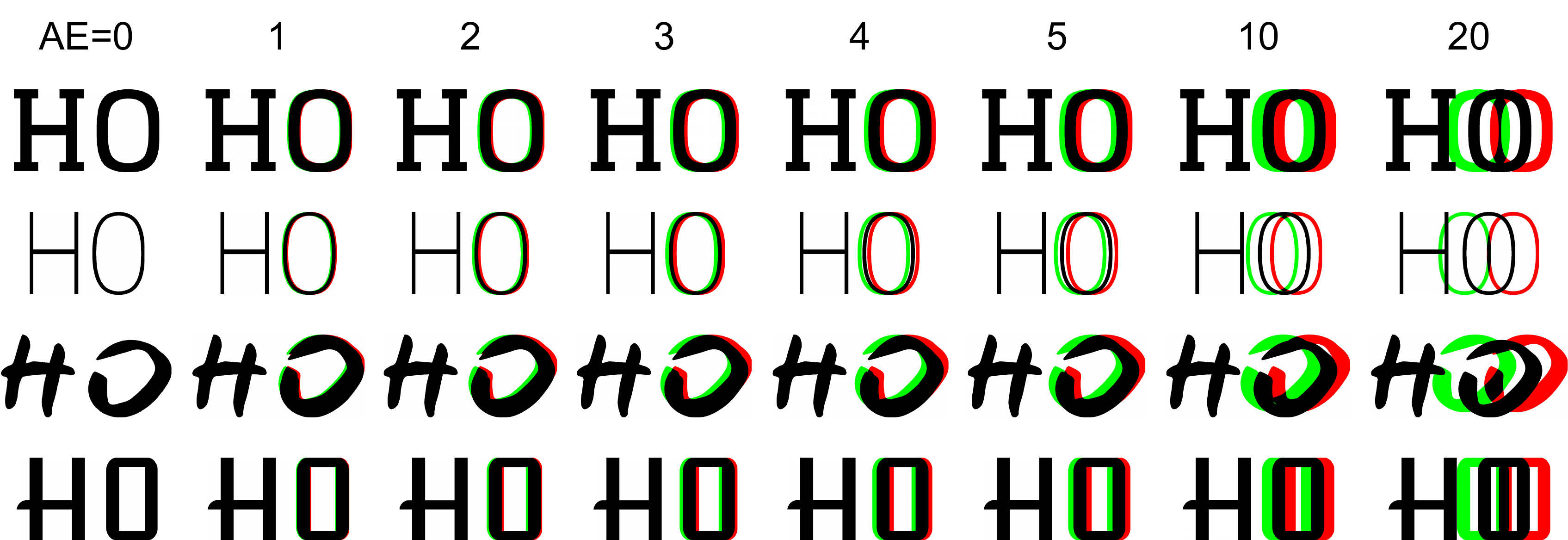}
    \caption{Letter pairs with different space estimation error values. (AE = Absolute error in pixels.) In each pair, the black letters are spaced with the ground-truth. The green and red letters are under-spaced and over-spaced by the specified AE value.}
    \label{fig:MAE_example}
\end{figure}
\begin{table}[t]
    \centering
    \caption{Comparison results on the test set ($N=52$, 256 fonts). See the main text for the details. ``HW'' means ``Handwriting.'' The 256 test fonts consist of 85 Serif, 115 Sans-serif, 14 Handwriting, and 42 Display fonts.}
    \label{tab:result_mae} 
    \begin{tabular}{r||r||r|r|r|r}
    MAE$\downarrow$   & \multicolumn{1}{|c||}{All} & \multicolumn{1}{c|}{Serif}          & \multicolumn{1}{c}{Sans.}      &  \multicolumn{1}{|c|}{HW}     & \multicolumn{1}{c}{Disp.}        \\  \hline
    Monospace                    & 22.472                         & 20.797                          & 20.369                               & 43.627                                & 24.572                            \\
    Average                      & 14.739                         & 11.213                          & 12.357                               & 41.148                                & 19.596                            \\ 
    FontForge                    & 9.857                         & 11.073                          & 8.560                               & \textbf{11.287}                                & 10.469                            \\ 
    pairwise (Ours)  & 13.271                         & 12.402                          & 12.068                               & 24.099                                & 14.714                            \\
    set-wise (Ours) & \textbf{5.318}                         & \textbf{5.224}                          & \textbf{3.996}                               & 11.366                                & \textbf{7.110}                            \\ 
    
    \multicolumn{6}{c}{}    \\
        \#fonts w/ MAE < 7$\uparrow$   & \multicolumn{1}{|c||}{All} & \multicolumn{1}{c|}{Serif}          & \multicolumn{1}{c}{Sans.}      &  \multicolumn{1}{|c|}{HW}     & \multicolumn{1}{c}{Disp.}        \\  \hline
    Monospace                    & 3                         & 0                          & 3                               & 0                                & 0                            \\
    Average                      & 70                         & 25                          & 41                               & 0                                & 4                            \\ 
    Fontforge                    & 59                         & 4                          & 46                               & \textbf{2}                                & 7                            \\ 
    pairwise  (Ours)          & 0                         & 0                          & 0                               & 0                                & 0                            \\
    set-wise   (Ours) & \textbf{203}                         & \textbf{72}                          & \textbf{105}                               & 0                                & \textbf{26}                            \\

    \multicolumn{6}{c}{}    \\

            \#wins$\uparrow$& \multicolumn{1}{|c||}{All} & \multicolumn{1}{c|}{Serif}          & \multicolumn{1}{c}{Sans.}      &  \multicolumn{1}{|c|}{HW}     & \multicolumn{1}{c}{Disp.}        \\  \hline
    Monospace                    & 3                         & 0                          & 3                               & 0                                & 0                            \\
    Average                      & 15                         & 10                          & 4                               & 0                                & 1                            \\ 
    FontForge                    & 21                         & 1                          & 6                               & \textbf{8}                                & 6                            \\ 
    pairwise  (Ours)              & 0                         & 0                          & 0                               & 0                                & 0                            \\
    set-wise (Ours) & \textbf{217}                         & \textbf{74}                          & \textbf{102}                               & 6                                & \textbf{35}                            \\
            
    \end{tabular}
\end{table}
\subsection{Comparative methods and evaluation metrics}
As noted before, there are no state-of-the-art machine learning-based letter space estimation models to be compared. As possible baselines under this situation, we prepare ``Monospace,'' ``Average,'' and ``FontForge.''  
\begin{itemize}
\item ``Monospace'' uses the constant letter space for all pairs from all fonts. For example, the pair \word{AB} and the pair \word{IJ} always have the same space regardless of font. The constant space is determined by averaging the ground-truth spaces of all letter pairs of all fonts in the training set.
\item ``Average'' determines the letter space between a certain letter pair (say, \word{AB}) by averaging the ground-truth spaces of that pair (\word{AB}) from all fonts in the training set. Consequently, the pair \word{AB} always has the same space regardless of font, whereas the spaces for \word{AB} and \word{IJ} are different. Although this is not a standard way for letter spacing, we introduce this to understand the importance of the letter space variations in different fonts. 
\item ``FontForge'' determines the letter space by the automatic kerning mode in FontForge, which is a {\it de-facto} standard software of font design. As noted before, it is based on a heuristic called optical spacing.
\end{itemize}
\par
We mainly use Absolute Error (AE) and MAE for the evaluation metric, which is also used as the loss function for the two models. AE indicates how much the ground-truth and the estimated space differ in the horizontal direction in pixels (when the image size is $256\times 256$). Fig.~\ref{fig:MAE_example} shows several examples of different AE values, visualizing two cases of the ground-truth $\pm$ AE with green and red letters. We also use a couple of other metrics (such as \#wins), which are detailed later.

\begin{figure}[t]
    \centering
    \includegraphics[width=.85\linewidth]{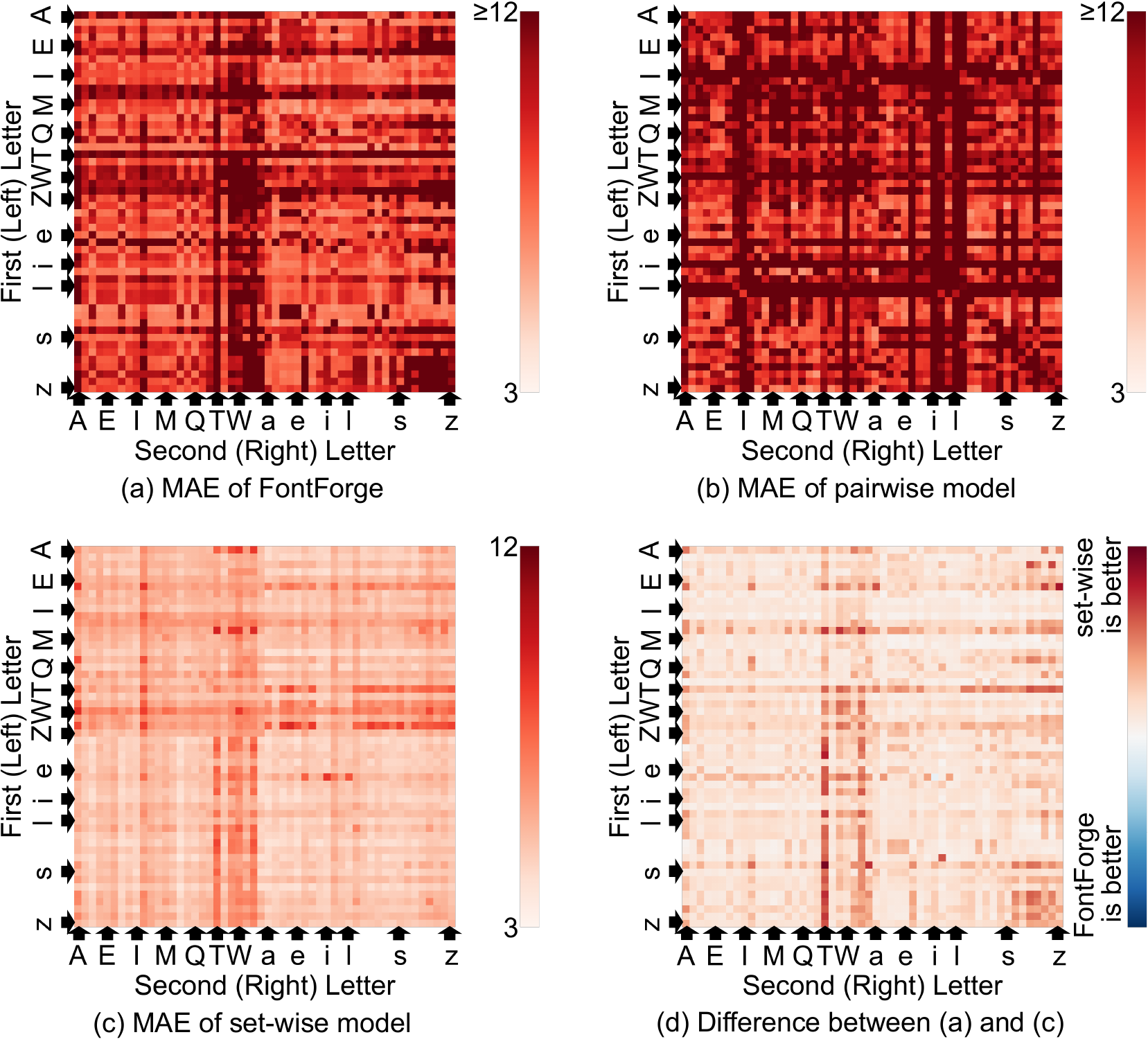}\\[-3mm] 
    \caption{(a)~MAEs of the FontForge. MAEs of the (b)~pairwise and (c)~set-wise models for all letter pairs. To observe small differences, MAEs of more than $12$ are shown with the same color as $12$ in (a) and (b). (d)~MAE difference between set-wise model and FontForge.} 
    \label{fig:result_compare_quality}  
\end{figure}
\begin{figure}[t]
    \centering
    \includegraphics[width=.8\linewidth]{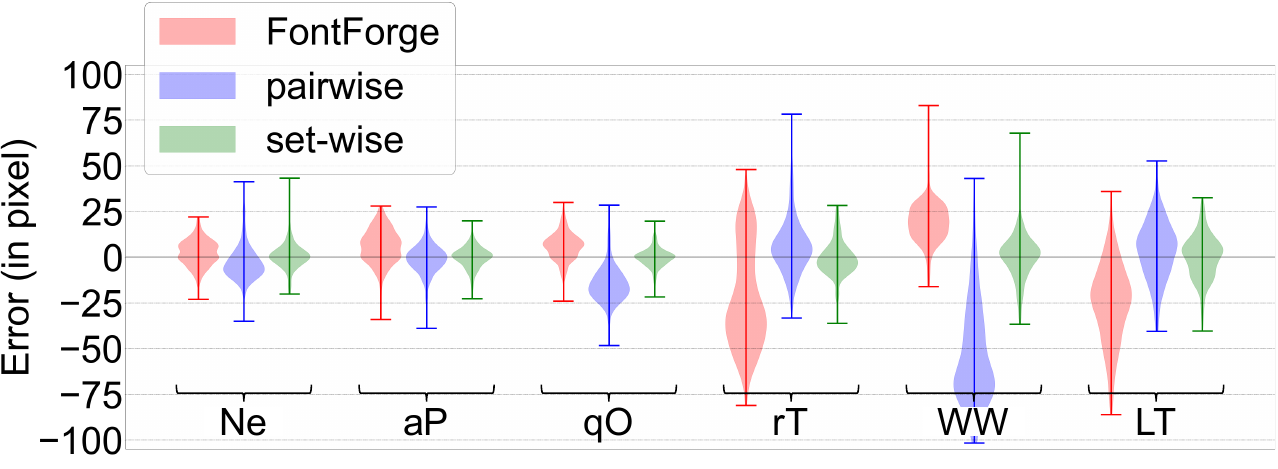}
    \caption{Estimation error distributions for several representative letter pairs. Specifically, FontForge, the pairwise model, and the set-wise model achieved the least errors for \wordc{Ne} \wordc{aP} and \wordc{qO} respectively and the largest error for \wordc{rT} \wordc{WW} and \wordc{LT} respectively.}
    \label{fig:violin}
\end{figure}
\begin{figure}[t]
    \centering
    \includegraphics[width=.8\linewidth]{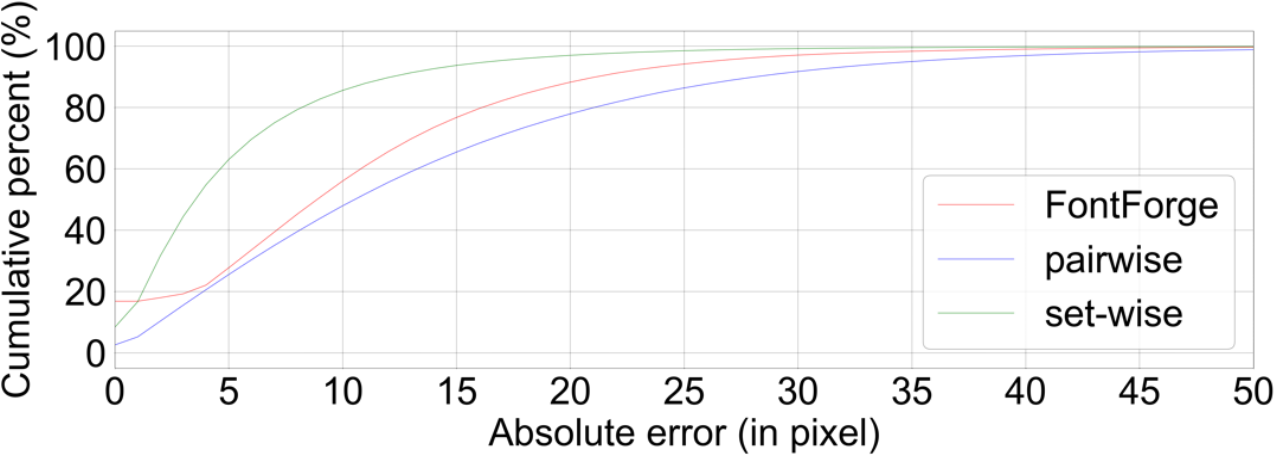}
    \caption{Cumulative percentage of the number of letter pairs that are below a specific absolute error. For example, the percentages of pairs with less than 10-pixel errors are 56\%, 48\%, and 86\% for FontForge, the pairwise model, and the set-wise model, respectively.}
    \label{fig:cumulative_percentage}
\end{figure}
\begin{figure}[t]
    \centering
    \includegraphics[width=\textwidth]{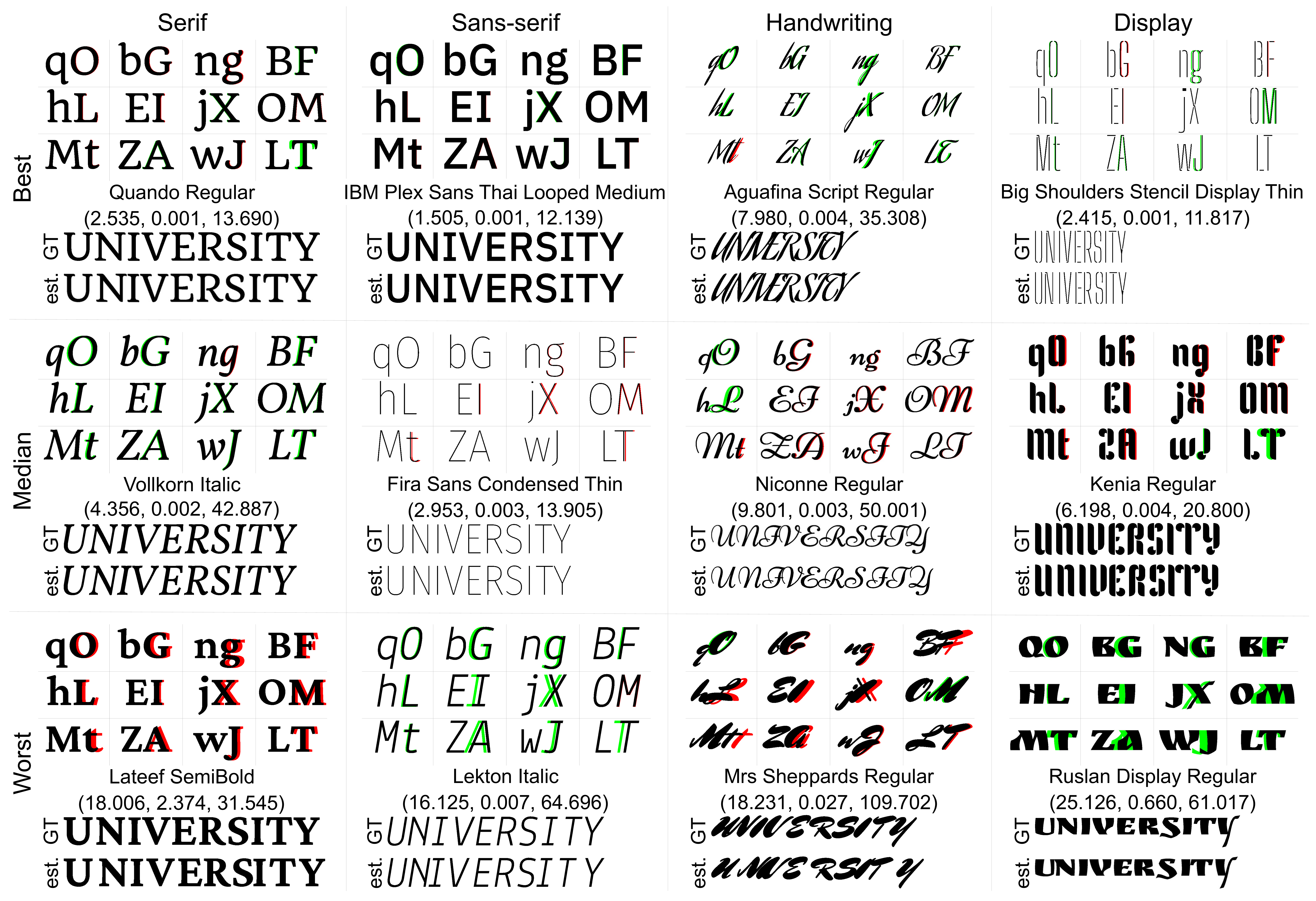} 
    \caption{The space estimation results by the set-wise model. For each font, twelve letter pairs (\wordc{qO} \wordc{bG}$\ldots$\wordc{wJ}\word{LT}) are shown in the visualization scheme of Fig.~\ref{fig:MAE_example}.
    We select three fonts with the smallest, median, and largest MAE in each style.
    Three parenthesized numbers are MAE, the minimum (i.e., best) AE, and the maximum (i.e., worst) AE in the $52\times 52$ letter spaces of the font. The word \word{UNIVERSITY} is printed with the ground-truth and estimated spaces.
    \label{fig:result_qualitative}} 
\end{figure}
\begin{figure}[t]
    \centering
    \includegraphics[width=\textwidth]{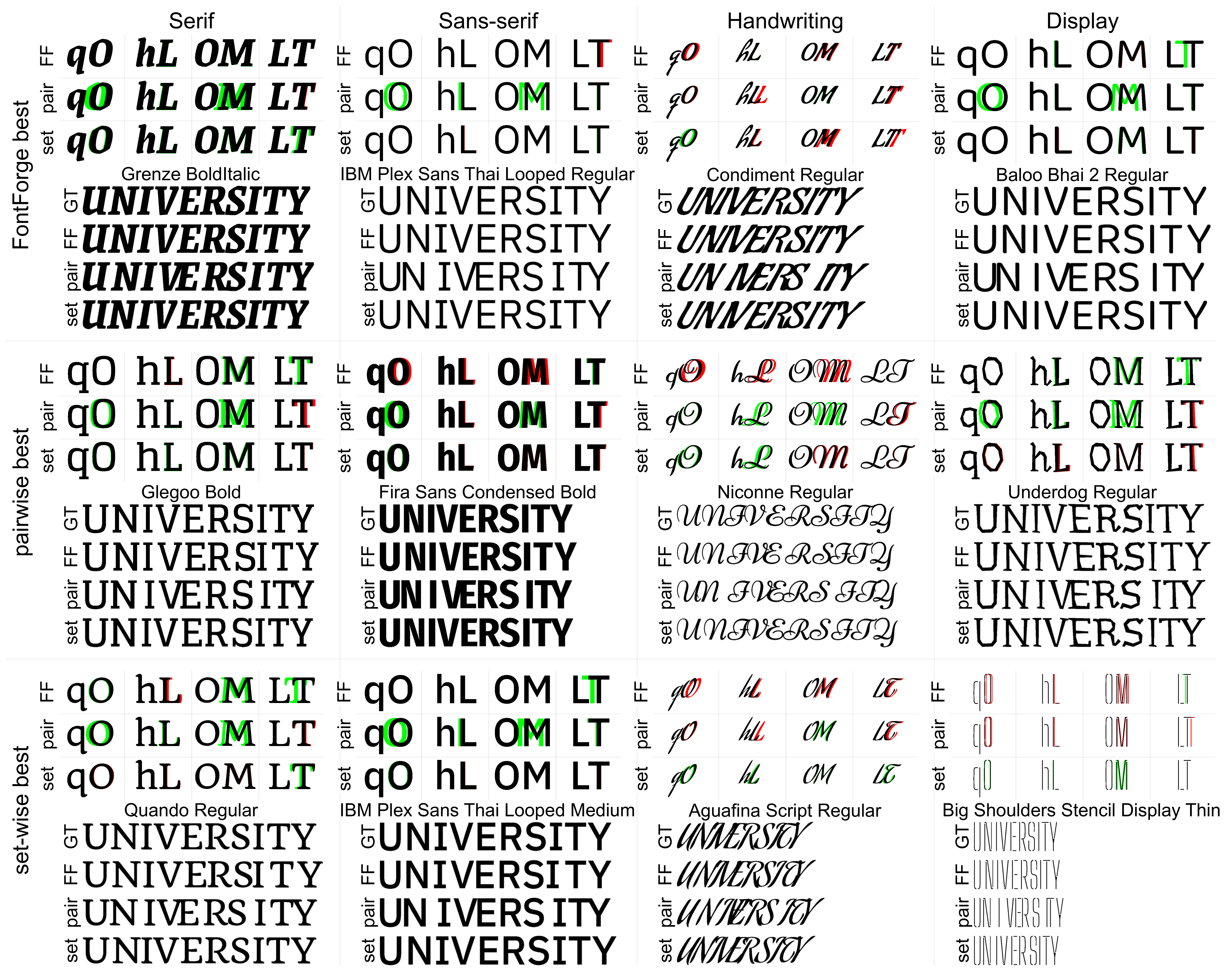}
    \caption{The best space estimation results by FontForge, the pairwise model, and the set-wise model, at each of four styles.  ``FF'' stands for FontForge.
    \label{fig:result_qualitative_c}} 
\end{figure}

\subsection{Quantitative evaluations}
Table~\ref{tab:result_mae} shows three quantitative evaluation results with different metrics. The first table shows the MAE values;  the ``All'' column shows the MAE of all $52\times 52$ letter pairs of all 256 test fonts, whereas the other columns show the MAE of the fonts in the corresponding style. This table proves that the set-wise model performs much better than not only the ``Monospace'' and ``Average'' methods but also the pairwise model. In addition, the set-wise model also outperforms  ``FontForge,'' except for a slight difference in the Handwriting style, where both of them show higher MAE values than the other styles. The superiority of the set-wise model indicates that considering all letters by the attention mechanism in Transformer is beneficial for the appropriate letter space estimation. 
\par
From the first table in Table~\ref{tab:result_mae}, we also observe that the set-wise model achieves an MAE of about $4.0$ pixels for Sans-serif fonts. This error is small enough, considering the average letter space is about 115 pixels. Another observation is that all the methods show larger MAE values for Handwriting fonts. One possible reason is that letters of Handwriting fonts often show large decorative strokes on their left and right sides, as we will see in our qualitative evaluation. Another reason is that the fonts with the Handwriting style are much fewer than those of the other styles and insufficient to cover large shape variations of the Handwriting fonts. (256 test fonts comprise 85 Serif, 115 Sans-serif, 14 Handwriting, and 42 Display fonts.) 
\par
The second (\#fonts with MAE < 7) and third (\#wins) tables in Table~\ref{tab:result_mae} still support the superiority of the set-wise model. The metric ``\#fonts with MAE <7'' is the number of fonts whose MAE (obtained for the $52\times 52$ letter pairs) is less than 7 pixels. (Note again the average letter space is 115 pixels --- therefore, the condition of ``MAE $< 7$'' is rather a hard requirement.) The metric ``\#wins'' is the number of fonts for which the method achieved the highest accuracy among the five methods. These detailed evaluations clearly show the superiority of the set-wise model over the other methods. Moreover, the second table shows that the set-wise model achieves MAEs of less than 7 pixels at 203 fonts out of 256.  
\par

Figs.~\ref{fig:result_compare_quality}(a), (b), and (c) visualize the MAE of each letter pair of all test fonts by FontForge, the pairwise model, and the set-wise model, respectively. Plot (c) is generally lighter than (a) and (b); this also indicates that the set-wise model achieved better performance for most pairs. The dark vertical and horizontal stripes on the plots (a)-(c) indicate the difficulty of estimating the space before and after a certain letter, respectively; for example, 
all those plots show dark stripes for \letter{W}, which as a large space variance in its ground-truth as shown in Fig.~\ref{fig:GT_Variance_withsmall}. A comparison between (b) and (c) clearly supports the superiority of the set-wise model over the pairwise model. Especially, the pairwise model could not give accurate estimations around thin letters, such as \letterc{I} \letterc{i} and \letterc{l} but the set-wise model could. As shown in Fig.~\ref{fig:GT_Variance_withsmall}, these thin letters have a deviated mean space from other letters, and the pairwise model could not catch the deviation correctly.
\par  
Fig.~\ref{fig:result_compare_quality}(d) shows the difference between (a) and (c). This plot indicates that the set-wise model shows better accuracy for most letter pairs than FontForge. Especially, the estimation accuracy around \letter{T} shows a large improvement. Although \letter{T} seems to be one of the ``notorious'' letters for letter spacing, the set-wise model could minimize the error.\par 

Fig.~\ref{fig:violin} shows the error distributions of the 256 test fonts for six letter pairs. \wordc{Ne} \wordc{aP} and \word{qO} are the easiest pairs with the least MAEs by FontForge, the pairwise model, and the set-wise models, respectively. In contrast, \wordc{rT} \wordc{WW} and \word{LT} are the worst pairs for them, respectively. The highlight of this plot is that the set-wise model still gives errors near zero
even in its worst pair (\word{LT}), whereas the other methods do not. Another highlight is that the set-wise model has lower error than FontForge, even on the best-performance pair for FontForge, \wordp{Ne}
 \par 
Fig.~\ref{fig:cumulative_percentage} shows the cumulative percentage of the letter pairs with less than a specific absolute error. (Here, the total number of the letter pairs is $256 \times 52 \times 52 $.) This graph also confirms the better performance of the set-wise model. In the set-wise model, 86\% of the pairs have an error of less than 10 pixels. In contrast, in the other two models, only around 50\% of pairs have less than a 10-pixel error. Note that FontForge achieved a 0-pixel error on 20\% of the pairs; this fact indicates that FontForge may have been used to set the default letter space of these pairs as the {\it de-facto} standard software.\par
\subsection{Qualitative evaluation\label{sec:qualitative}}
Fig.~\ref{fig:result_qualitative} shows 12 letter pairs printed in 12 different fonts with the letter spaces estimated by the set-wise model. The 12 letter pairs (\wordc{qO} \wordc{bG}$\ldots$, \wordc{wJ} \word{LT}) were picked up at equal rank intervals from the best pair (\word{qO}) with the smallest MAE to the worst pair (\word{LT}). For each of the four font styles, we select three fonts that give the best (i.e., the smallest), median, and worst MAEs, respectively. \par
These results, especially those with small MAEs (less than about two pixels), show that the set-wise model performs near-perfect space estimation. More importantly, the results are still acceptable even when larger MAEs. In fact, in Fig.~\ref{fig:result_qualitative}, the word \word{UNIVERSITY} with the estimated spaces has no extra space to split the word into fragments and no under-estimated space to overlap adjacent letters, except for {\tt Mrs Sheppards Regular}, where letters in a Handwriting style are overlapping. Another interesting result is {\tt Lekton Italic}, which becomes the worst due to the under-estimated spaces of several pairs; however, the under-estimated spaces still seem also (or even more) natural.\par 

Fig.~\ref{fig:result_qualitative_c} shows the best cases of the three methods, FontForge, the pairwise model, and the set-wise model, at each of four font styles. The four letter pairs, \wordc{qO} \wordc{hL} \wordc{OM} and \wordc{LT} are picked from the pairs in 
Fig.~\ref{fig:result_qualitative}. For the comparison among the methods, these letter pairs and the word \word{UNIVERSITY} are printed with the letter space estimated by all methods for each of the 12 fonts. The letter pairs show that even for fonts where FontForge or the pairwise model give the best results, the set-wise model shows similar accuracy. In contrast, for the fonts where the set-wise model gives the best, FontForge and the pairwise model sometimes show a large deviation from the ground-truth. As also shown in Fig.~\ref{fig:violin}, this observation proves the set-wise model is more stable than the others. Note that the word images \word{UNIVERSITY} by FontForge often show differences from the ground-truth; this means that FontForge is insufficient for the automatic letter space estimation and its results need manual corrections.\par

\begin{figure}[t]
\begin{tabular}{cc}
  \begin{minipage}{.30\textwidth}
    \centering
    \includegraphics[width=.9\linewidth]{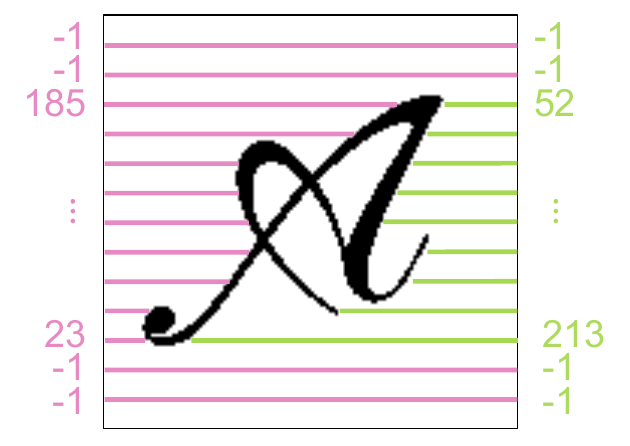}\\[-3mm]
    \figcaption{Peripheral feature.}
    \label{fig:peripheral}
  \end{minipage}
  \hspace{0.5cm}
 \begin{minipage}{.65\textwidth}
    \begin{center}
    \tblcaption{Comparison between ResNet-based feature and peripheral feature in the same set-wise model architecture. The former is used in all previous experiments, and therefore, its result is the same as Table~\ref{tab:result_mae}.
    \label{tab:compare_peripheral}}
    \vspace{3mm}
    \begin{tabular}{c||r||r|r|r|r}
    MAE$\downarrow$   & \multicolumn{1}{|c||}{All} & \multicolumn{1}{c|}{Serif}          & \multicolumn{1}{c}{Sans.}      &  \multicolumn{1}{|c|}{HW}     & \multicolumn{1}{c}{Disp.}        \\  \hline
    ResNet feature & \textbf{5.318}                         & \textbf{5.224}                          & \textbf{3.996}                               & \textbf{11.366}                                & \textbf{7.110}                            \\     
    Peripheral     & 10.424                         & 10.029                          & 7.687                               & 26.148                                & 13.480                            \\ 
    \end{tabular}
    \end{center}
  \end{minipage}
  \end{tabular}
\end{figure}

\section{Letter Space Estimation by Peripheral Feature\label{sec:peripheral}}
Another letter space estimation experiment is performed with the peripheral feature of Fig.~\ref{fig:peripheral}. The 512-dimensional ResNet feature used in the above experiments represents the whole letter shape. The peripheral feature, in contrast, does not represent the whole letter shape; it represents the shapes of the left and right sides of a letter. More specifically, it is a 512-dimensional feature for a $256\times 256$ letter image, and its element is the distance between the right edge of the image and the right edge of the letter stroke or between the left edge of the image and the left edge of the letter stroke at each height.
\par
As noted in Section~\ref{sec:intro}, the conventional kerning process often depends on {\em optical spacing}, which is based on the blank space area between adjacent letters. This means the conventional process does consider the peripheral features (representing the blank space) rather than the whole letter shape. Therefore, this experiment will reveal whether the optical spacing is a sufficient heuristic feature for kerning. Note that the experiment was performed by using the set-wise model with the same Transformer architecture trained by the same procedure as the previous experiments; namely, the difference is the input feature only.
\par

Table~\ref{tab:compare_peripheral} shows MAE for all 256 test fonts and  MAEs for individual styles. (Note again that the set-wise models give these MAEs.) Surprisingly, there are large differences between the performance of the peripheral feature and the ResNet feature. (Recall that ResNet for the feature extraction is trained on a letter classification task and then frozen; that is, it is not fine-tuned for kerning.) These performance differences prove that using the whole letter shape is more beneficial than the side shapes. The conventional heuristic process of optical spacing might be effective for rough kerning but insufficient for more precise kerning.
\section{Conclusion, Limitation, and Future Work}
This study is the first attempt to estimate optimal letter space by machine learning. We proposed two models, pairwise and set-wise models. The pairwise model is based on a deep neural network (DNN) and accepts two letter images (say, \letter{A} and \letter{B}) and estimates their appropriate space.
In contrast, the set-wise model is based on Transformer and accepts $N$ letter images (say, \letterc{A} $\ldots$ \letter{Z}) of a certain font as its inputs and estimates the letter spaces for each of $N^2$ letter pairs simultaneously. These models are trained using existing fonts whose default letter spaces are pre-determined by human experts. \par
Experimental results on 2558 Google fonts show that the set-wise model can give accurate letter spaces. More specifically, when the average letter space is about 115, the set-wise model's mean absolute error (MAE) is 5.318 pixels for the 52 Latin alphabet letters from the 256 fonts in our test set. The set-wise model showed better MAE on 217 among 256 fonts than the pairwise model and three baselines, including FontForge. This superiority comes from the self-attention mechanism of Transformer, by which space estimation for specific two letters (say, \letter{A} and \letter{B}) is helped by all the other letters (\letter{C} to \letter{Z} and \letter{a} to \letter{z}). This is reasonable because kerning for a certain font should be done carefully by balancing the spaces of all letter pairs. Through various quantitative and qualitative analyses of the estimation results, we clarified when and how estimation accuracy is affected by letter classes and the models. We also showed that the letter feature extracted from the whole letter shape was better than a peripheral feature.\par
Our current limitation is the inflexibility of the estimation output. In this paper, we treated the default letter spaces as the ground-truth and tried to estimate the letter spaces similar to the ground-truth. In other words, we assumed that the appropriate letter space is uniquely determined for each letter pair of each font. However, typographic designers modify the letter spaces to create a particular impression. For example, letter spacing slightly wider than the default gives a more established and confident impression. In future work, such a space modification mechanism should be introduced into the current model to allow for more flexible kerning. We plan to apply our model to AI-generated fonts and expand our model to estimate letter spaces while considering three or more letters or even a paragraph (like the hz-program). In addition, we will modify our model to deal with font data in vector formats, although we used raster images. Finally, subjective evaluation by the kerning experts is mandatory.\par

\par
\noindent{\bf Acknowledgment}:\ This work was supported by JSPS KAKENHI Grant Number JP22H00540. We thank anonymous reviewers for their professional knowledge about kerning.
%
\bibliographystyle{splncs04}
\bibliography{sample-base}

\end{document}